\documentclass[conference]{IEEEtran}
\IEEEoverridecommandlockouts

\usepackage{cite}
\usepackage{amsmath,amssymb,amsfonts}
\usepackage{booktabs} 
\usepackage{multirow} 
\usepackage{graphicx} 
\usepackage{array} 
\usepackage{algpseudocode}
\usepackage{amsmath}  
\usepackage{caption}  
\usepackage{algorithm}
\usepackage{subfigure}
\usepackage{graphicx}
\usepackage{textcomp}
\usepackage{xcolor}
\def\BibTeX{{\rm B\kern-.05em{\sc i\kern-.025em b}\kern-.08em
    T\kern-.1667em\lower.7ex\hbox{E}\kern-.125emX}}

\begin{document}

\title{LoR\textsuperscript{2}C : Low-Rank Residual Connection Adaptation for Parameter-Efficient Fine-Tuning}

\author{
\IEEEauthorblockN{
    Jiancheng Zhao$^{\star}$, 
    Xingda Yu$^{\star}$, 
    Yuxiang Zhang, 
    Zhen Yang$^{\ast}$
}
\IEEEauthorblockA{
    \textit{School of Computer Science and Technology} \\
    \textit{Shandong University, Qingdao, China} \\
    \{202200120166, 202200120014, zhangyuxiang1412\}@mail.sdu.edu.cn, zhenyang@sdu.edu.cn
}
\thanks{$^{\star}$ These authors contributed equally to this work.}
\thanks{$^{\ast}$ Corresponding author.}
}

\maketitle

\begin{abstract}

In recent years, pretrained large language models have demonstrated outstanding performance across various natural language processing tasks. However, full-parameter fine-tuning methods require adjusting all model parameters, leading to immense computational resource demands. Although parameter-efficient fine-tuning methods like LoRA have significantly reduced the number of parameters, they still face challenges such as gradient vanishing and the potential for further parameter reduction.
To address these issues, this paper proposes a novel parameter-efficient fine-tuning method called LoR\textsuperscript{2}C (Low-Rank Residual Connection Adaptation). LoR\textsuperscript{2}C introduces residual connections with low-rank matrices within the model layers, which not only reduces the number of fine-tuning parameters but also effectively alleviates the gradient vanishing problem. Additionally, this paper presents three optimization variants of LoR\textsuperscript{2}C: ShareLoR\textsuperscript{2}C, MergeLoR\textsuperscript{2}C, and InjectLoR\textsuperscript{2}C. These variants further improve parameter efficiency and model performance through parameter sharing, module merging, and injection mechanisms, respectively.
Experimental results on multiple natural language understanding and natural language generation tasks demonstrate that LoR\textsuperscript{2}C and its optimized variants significantly reduce parameter overhead while maintaining or even improving performance, outperforming existing mainstream parameter-efficient fine-tuning methods.Our code is publicly available at https://github.com/Oblivioniss/LoR2C.
\end{abstract}


\section{Introduction}
In recent years, the scale of large language models (LLM) has grown rapidly and these models have demonstrated exceptional performance on various tasks. However, despite the significant performance improvements that full parameter fine-tuning (FT) can bring, adjusting all the model parameters not only consumes massive computational resources, but also may lead to overfitting and inefficient training.

To address these challenges, researchers have proposed Parameter-Efficient Fine-Tuning (PEFT) methods aimed at reducing computational costs while maintaining fine-tuning effectiveness. LoRA \cite{b1} emerged in this context. LoRA reduces computational and storage overhead by adjusting only a subset of the model parameters while still achieving significant fine-tuning improvements without compromising performance.
\begin{figure}[t]
\centering
\subfigure[]
{
    \begin{minipage}[b]{.45\linewidth}
        \centering
        \includegraphics[scale=0.24]{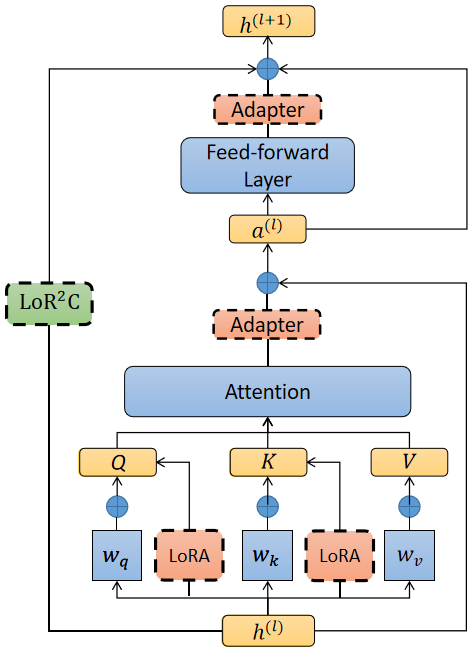}
    \end{minipage}
}
\subfigure[]
{
 	\begin{minipage}[b]{.45\linewidth}
        \centering
        \includegraphics[scale=0.24]{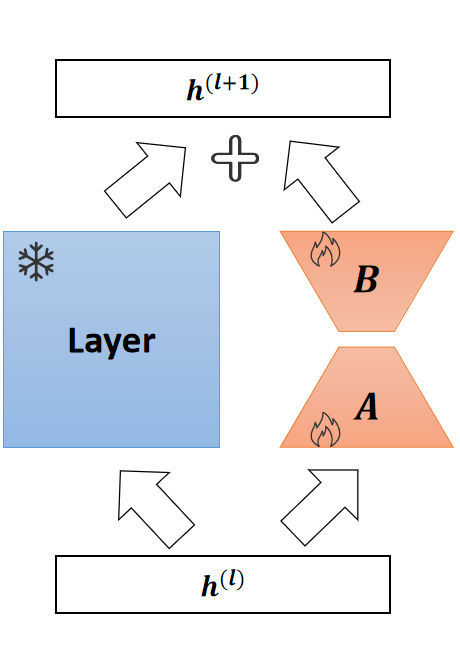}
    \end{minipage}
}

\caption{Explanation of LoR\textsuperscript{2}C:
(a) The figure illustrates the differences in positioning between LoR\textsuperscript{2}C, LoRA, and Adapter. LoR\textsuperscript{2}C introduces an additional residual connection within the Transformer layers to enhance gradient propagation, effectively mitigating the gradient vanishing problem. 
(b) The figure provides a detailed explanation of the low-rank decomposition process of matrix W in the LoR\textsuperscript{2}C module, where W=BA. By decomposing W into two low-rank matrices A and B, LoR\textsuperscript{2}C further reduces the parameter count.}
\end{figure}
Although LoRA has made initial strides, its basic form still has some limitations. To overcome these issues, new improvements have emerged. For instance, MeLoRA \cite{b2} connects multiple small LoRA modules in parallel, ensuring a higher rank while reducing the number of parameters, thus improving performance. ResLoRA \cite{b3} introduces residual paths during training to accelerate gradient propagation, addressing the problem of gradient vanishing in LoRA. However, these existing methods still fail to simultaneously address both the problem of excessive parameters and the gradient vanishing issue effectively.

Residual connections in ResNet \cite{b4}, as an effective structure, have been applied to the self-attention and feed-forward layers of Transformers \cite{b5}. Inspired by this, we propose a fine-tuning method called LoR\textsuperscript{2}C (Low Rank Residual Connection Adaptation). As shown in Fig. 1, in our approach, we introduce residual connections with matrix transformations across the entire layer. Experimental results show that the transformation matrix exhibits low-rank properties. Based on the LoRA principle, we replace the transformation matrix with the product of matrices A and B.

To further reduce the parameter count and fully utilize the available parameters, we also investigated variants of LoR\textsuperscript{2}C. Drawing inspiration from the ShareLoRA \cite{b6} concept, we propose a strategy where matrix A is shared across all LoR\textsuperscript{2}C models, while matrix B remains independent, thus reducing the overall parameter count. In addition, we introduce two mechanisms: the merging mechanism and the injection mechanism. These mechanisms are independent, but can be applied concurrently.

In the injection mechanism, we partition the training process into three distinct phases. First, during the initialization phase, LoR\textsuperscript{2}C is introduced at each layer to capture layer-specific information. In the subsequent injection phase, we employ the Shape of Feature Space (SFS) to assess the information content of each LoR\textsuperscript{2}C matrix. Matrices with the lowest information content are then substituted with low-rank LoRA matrices. This iterative process continues until the final network structure is determined, at which point the model undergoes multiple rounds of training to ensure convergence. 

The merging mechanism operates similarly to the injection mechanism, but with a distinction in the merging phase. In this phase, we evaluate the combined information content of adjacent matrices by summing their SFS values. The two LoR\textsuperscript{2}C matrices with the smallest cumulative information are fused into a single LoR\textsuperscript{2}C that spans two layers, thus further reducing the parameter count while preserving the model’s performance.

Compared to LoRA, our method is not only parameter-efficient but also mitigates the gradient vanishing issue through residual connections. We conduct extensive experiments across multiple tasks and models, including natural language understanding (NLU) tasks on the GLUE \cite{b7} dataset using RoBERTa-base \cite{b8} and instruction fine-tuning tasks on LLAMA2-7B \cite{b9}. The experimental results demonstrate the effectiveness of our method. Overall, our contributions are as follows:

\begin{itemize} \item We propose a novel fine-tuning method, LoR\textsuperscript{2}C, which alleviates the gradient vanishing issue with fewer parameters.\item We introduce and explore LoR\textsuperscript{2}C-based optimizations, namely parameter-sharing LoR\textsuperscript{2}C, and LoR\textsuperscript{2}C with merging and injection mechanisms. These methods can be used independently or together. \item Through experiments, we compare LoR\textsuperscript{2}C and its optimizations with existing methods across multiple tasks. The results show that LoR\textsuperscript{2}C achieves higher parameter efficiency and superior performance. \end{itemize}
\section{Related Work}
\subsection{Parameter-Efficient Fine-Tuning}
PEFT methods aim to reduce the adjustments required for large-scale pre-trained models during fine-tuning by optimizing only a small subset of parameters. This significantly reduces computational and storage costs. The core idea of these methods is to keep most of the pre-trained model parameters fixed while adding a small number of task-specific parameters to improve fine-tuning efficiency. Common PEFT methods include Adapter\cite{b10}, LoRA, and Prefix Tuning \cite{b11}.The Adapter method inserts small additional adapter layers in each layer, which are updated during fine-tuning while other original parameters remain frozen. This approach significantly reduces the number of parameters required for fine-tuning. By sharing adapter parameters across tasks, the method improves parameter efficiency in multi-task learning.Prefix Tuning adapts models by optimizing task-specific prefix vectors while keeping the model's parameters frozen. This method typically adjusts fewer parameters and is particularly suitable for generative tasks.They are often used in multiple fields, such as Automated Code Translation \cite{b99}.Although these PEFT methods effectively reduce computational demands, they still face challenges in balancing parameter efficiency and performance in large-scale models. As a result, more efficient fine-tuning methods remain an active area of research.
\subsection{LoRA (Low-Rank Adaptation)}
Recently, researchers have proposed several improvements to LoRA. For example, DyLoRA\cite{b12} reduces the parameter usage of unimportant layers by dynamically adjusting the rank for each layer. AsLoRA \cite{b13} further reduces parameters by adaptively merging B based on ShareLoRA. AdaLoRA\cite{b14} dynamically adjusts the rank based on the importance of each layer, further lowering computational costs. Additionally, the combination of LoRA with other techniques has also gained attention, including AdaMix\cite{b15} and QLoRA\cite{b16}, which optimize model performance and reduce storage demands by adjusting module parameters and applying quantization. HydraLoRA \cite{b17} and LoRAMoE \cite{b18} explore the use of LoRA modules in multi-task learning scenarios. Although LoRA significantly reduces the number of parameters, gradient vanishing issues may still occur in deep networks. To address this, ResLoRA introduces residual paths to mitigate the gradient vanishing problem. Our approach further tackles both the gradient vanishing issue and the reduction of parameters.
\subsection{ResNet (Residual Networks)}
ResNet is a structure that optimizes the training of deep networks by introducing residual connections (skip connections). The key advantage of ResNet lies in its ability to effectively alleviate the gradient vanishing problem in deep neural networks, thereby enhancing gradient propagation and accelerating network convergence. The design principles of ResNet have been widely applied to various network structures, including the self-attention and feed-forward layers of Transformer models. Inspired by ResNet, ResLoRA introduces residual paths to improve the training performance of LoRA in deep networks and to mitigate the gradient vanishing problem. Unlike ResLoRA, our method combines residual connections with low-rank matrix transformations to further reduce the parameter count.

\begin{figure}[t]
    \centering
    \includegraphics[scale=0.2]{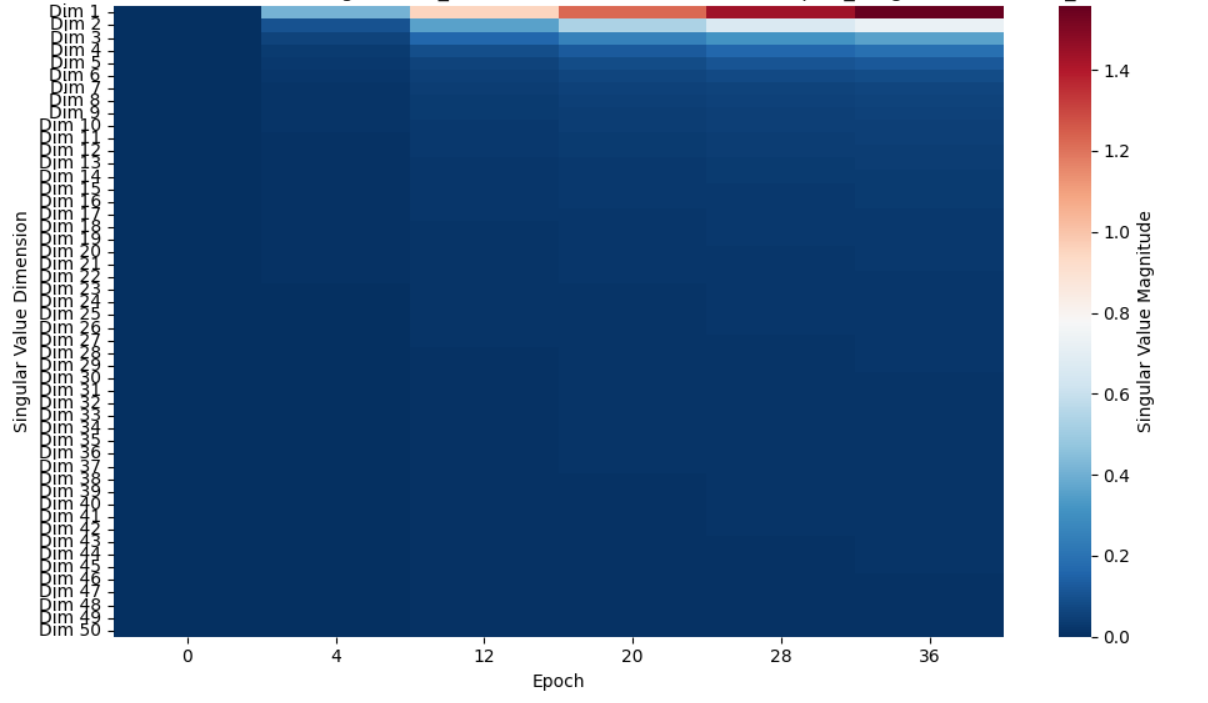} 
    \caption{Heatmap of the top 50 singular values of the \( W \) matrix, averaged across all 12 layers during training on the MRPC dataset. 
The vertical axis represents the index of singular values sorted in descending order, and the horizontal axis represents the training epochs (up to 50). 
The dimension of the \( W \) matrix is \(\text{in\_features} \times 128\) (where \(\text{in\_features} = 768\)). 
This experiment aims to explore whether the \( W \) matrix exhibits low-rank properties during training. 
The heatmap shows that only the first few singular values have significant magnitudes, while the majority diminish rapidly, indicating that the \( W \) matrix retains low-rank characteristics. 
Training settings include a batch size of 64, a maximum sequence length of 256, and a learning rate of \(4 \times 10^{-4}\). 
}
    \label{fig:example}
\end{figure}
\section{Method}
In this section, we present our framework, which consists of three components: (1) LoR\textsuperscript{2}C, the basic method that introduces residual connections to layers and is used for fine-tuning; (2) ShareLoR\textsuperscript{2}C, an extension of LoR\textsuperscript{2}C that shares parameters across layers; and (3) MergeLoR\textsuperscript{2}C and InjectLoR\textsuperscript{2}C, LoR\textsuperscript{2}C variants based on merging and injection mechanisms.
\subsection{Preliminaries on Low-Rank Adapter}
We begin by introducing LoR\textsuperscript{2}C in the context of other fine-tuning methods.  
As shown in Fig. 1, the LoRA module is commonly applied to the attention mechanism in Transformer architectures. Specifically, LoRA is typically integrated into the linear projection weight matrices for Query (\( W_Q \)) and Value (\( W_V \)), while the rest of the model parameters remain frozen. In the attention layer of a Transformer, the core operation involves mapping input features to the Query (\( Q \)), Key (\( K \)), and Value (\( V \)) spaces using weight matrices. To reduce the parameter overhead during fine-tuning, LoRA introduces two low-rank matrices \( A \) and \( B \), where \( A \in \mathbb{R}^{ r\times d} \) and \( B \in \mathbb{R}^{d \times r} \) \( (r \ll d) \), to replace directly updating \( W_Q \) and \( W_V \). Specifically, the update rule for \( W_Q \) is as follows:
\begin{equation}
W_Q' = W_Q + \Delta W_Q = W_Q + BA
\end{equation}
where \( \Delta W_Q = BA \) represents the weight update term calculated using the low-rank matrices \( A \) and \( B \), and \( W_Q' \) is the updated Query weight matrix. Similarly, the update for \( W_V \) is also achieved through the product of low-rank matrices. This approach significantly reduces the number of parameters that need to be optimized, while retaining the knowledge encoded in the pre-trained weights, thereby improving fine-tuning efficiency and computational effectiveness.

\begin{figure}[t]
    \centering
    \includegraphics[scale=0.5]{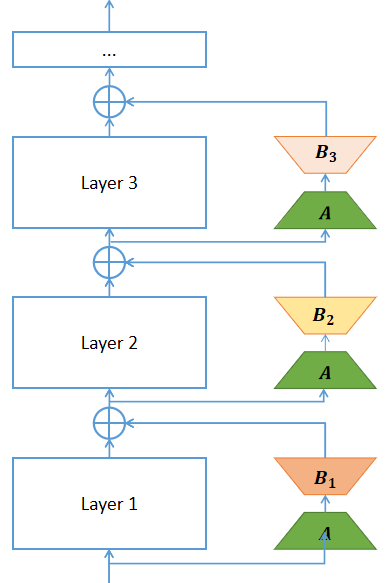} 
    \caption{An illustration of the architecture of ShareLoR\textsuperscript{2}C. The figure demonstrates how the matrix \( A \) is shared across all layers, while each layer retains its own independent matrix \( B_t \) (e.g., \( B_1, B_2, B_3 \)).
}
    \label{fig:example}
\end{figure}

\begin{figure*}[t]
    \centering
    \includegraphics[scale=0.45]{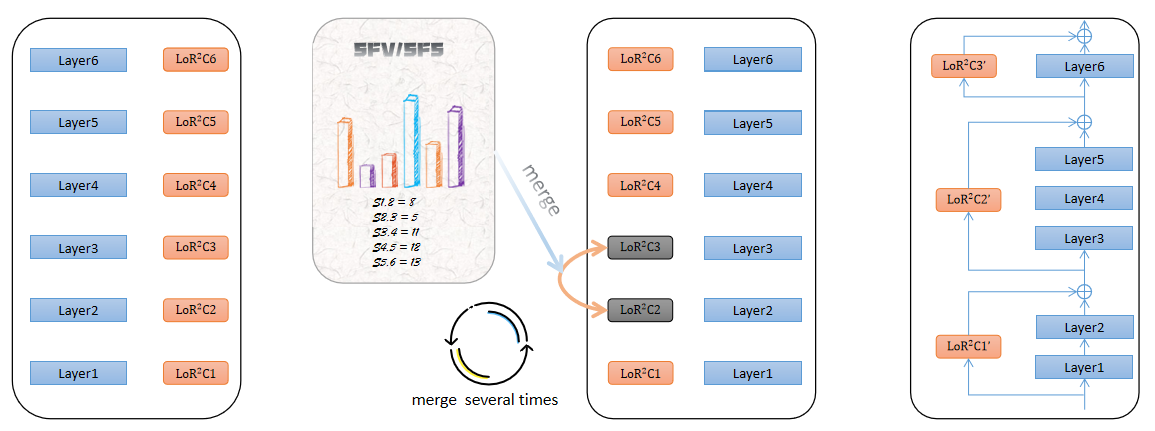} 
    \caption{This figure illustrates the MergeLoR\textsuperscript{2}C mechanism. In the initial structure (leftmost panel), each layer contains a LoR\textsuperscript{2}C module (\( \text{LoR}^2\text{C}_1, \text{LoR}^2\text{C}_2, \dots, \text{LoR}^2\text{C}_6 \)). The SFS metricss are used to evaluate the information content of each module (second panel). Layers with the lowest combined information scores are identified, and their corresponding LoR\textsuperscript{2}C modules are merged into a single module (third panel). This process is repeated multiple times, progressively reducing the number of LoR\textsuperscript{2}C modules. The final architecture (rightmost panel) incorporates merged modules (\( \text{LoR}^2\text{C}_1', \text{LoR}^2\text{C}_2', \text{LoR}^2\text{C}_3' \)) across multiple layers, along with residual connections to preserve gradient flow and ensure parameter-efficient.
}
    \label{fig:example}
\end{figure*}

\subsection{Low Rank Residual Connection Adaptation}
To mitigate the gradient vanishing problem, we introduce residual connections across the entire layer, where the residual connections include a trainable matrix \( W \). Specifically, the input from the previous layer, \( h(t) \), is multiplied by the matrix \( W \) and then added to the input of the current layer through the residual connection. The resulting output is passed to the next layer, which can be expressed as:
\begin{equation}
h(t+1) = h(t)W + \text{Layer}(h(t))
\end{equation}
Inspired by the LoRA approach, we follow a similar methodology, as demonstrated in the LoRA paper, and experimentally verify the low-rank property of matrix \( W \)(Fig. 2). Therefore, we represent \( W \) as \( W = BA \), where \( A \in \mathbb{R}^{r \times d} \) and \( B \in \mathbb{R}^{d \times r} \), with \( r \ll d \). Compared to LoRA, this method reduces the parameter count by half for matrices with the same rank.However, the introduction of additional computations in this method inevitably incurs some inference latency.
\subsection{ShareLoR\textsuperscript{2}C}
To further reduce the storage and computational overhead of the model, we  attempted ShareLoR\textsuperscript{2}C, which introduces a parameter-sharing mechanism based on LoR\textsuperscript{2}C.We show our structure in Fig. 3. In ShareLoR\textsuperscript{2}C, the matrix \( A \) is shared across different layers, while the matrix \( B \) remains independent. Specifically, the matrix \( A \) is shared across all layers, and each layer's matrix \( B \) is trained independently. By sharing matrix \( A \), we significantly reduce the number of parameters required for storage while preserving each layer's ability to adapt to specific tasks.

In ShareLoR\textsuperscript{2}C, the update formula is given by:
\begin{equation}
h(t+1) = h(t)B_tA + \text{Layer}(h(t))
\end{equation}
where \( t \) is the index of the layer in the model, \( A \) is a trainable matrix shared across all layers, and \( B_t \) represents the matrix \( B \) specific to layer \( t \).

During training, the matrix \( A \) is shared among all layers, so only one copy of \( A \) needs to be stored, while \( B_t \) is optimized independently for each layer. With this design, we aim to significantly reduce the storage cost of the model while retaining the specific representational capacity of each layer, thereby enhancing the model's performance on specific tasks.
\subsection{MergeLoR\textsuperscript{2}C\& InjectLoR\textsuperscript{2}C}

\textbf{MergeLoR\textsuperscript{2}C:} The overall workflow of the MergeLoR\textsuperscript{2}C method is illustrated in Fig. 4. This method is designed to reduce the number of parameters and alleviate inference latency by iteratively merging LoR\textsuperscript{2}C modules.
At the beginning of training, we initialize and train the LoR\textsuperscript{2}C modules in each layer to adapt to the specific task requirements. In each layer, the LoR\textsuperscript{2}C module updates weights by introducing low-rank matrices \( A_t \) and \( B_t \) as follows:
\begin{equation}
W_t' = W_t + B_tA_t
\end{equation}
where \( W_t \) represents the frozen pre-trained weights, and \( A_t \) and \( B_t \) are newly added trainable parameters. The training objective is to enable each LoR\textsuperscript{2}C module to effectively capture the information characteristics of its respective layer, which can then be used for the merging operation.

After a specific number of training iterations, we assess the information content of each \( W \) matrix (decomposed into \( A \) and \( B \)) by analyzing its singular values. Based on the size of the information content, merging operations are performed. We define the SFS to estimate the information content of matrices:

\begin{figure*}[t]
    \centering
    \includegraphics[scale=0.45]{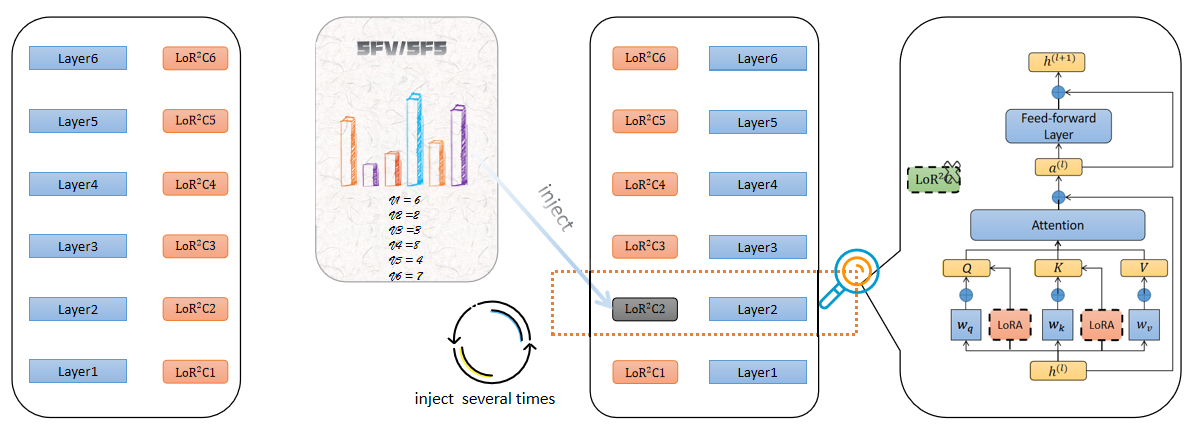} 
    \caption{This figure illustrates the InjectLoR\textsuperscript{2}C mechanism. In the initial structure (leftmost panel), each layer contains a LoR\textsuperscript{2}C module (\( \text{LoR}^2\text{C}_1, \text{LoR}^2\text{C}_2, \dots, \text{LoR}^2\text{C}_6 \)). The SFS metrics is used to evaluate the information content of each module (second panel). Based on the metric, the module with the lowest information score is identified, and its corresponding LoR\textsuperscript{2}C module is replaced by a lower-rank LoRA module (third panel). This injection process is repeated multiple times, progressively reducing the rank of selected modules. The final architecture (rightmost panel) integrates the injected LoRA modules into the Transformer layers, where LoRA modules are applied to the query (\( W_Q \)) and value (\( W_V \)) weight matrices.
}
    \label{fig:example}
\end{figure*}
 **SFS (Shape of Feature Space): 
   The SFS metrics analyzes the singular value distribution of the output feature matrix from the LoR\textsuperscript{2}C module to evaluate the structural characteristics of the feature space. Specifically, SFS computes the proportion of the sum of the remaining singular values (excluding the top \( k \)) relative to the total sum of singular values. The formula is given as:
\begin{equation}
   \text{SFS}(t) = 1 - \frac{\sum_{i=1}^{k} \lambda_i}{\sum_{j=1}^{n} \lambda_j}
\end{equation}
where $\lambda_i$ is the $i$-th singular value of matrix $W$ obtained through SVD decomposition arranged in descending order, \( k \) is the number of selected top singular values, and \( n \) is the total number of singular values. SFS effectively evaluates the concentration of information in a matrix by quantifying the proportion of the first 
k singular values relative to the total singular values. A low SFS value indicates that the information is concentrated in a few singular values, making the matrix suitable for low-rank approximation. In contrast, a high SFS value suggests that the information is more evenly distributed, requiring the retention of more singular values. 
By computing the SFS metrics, we can intuitively evaluate the importance of each module in terms of feature representation. Through the SFS metrics, we sort all LoR\textsuperscript{2}C modules in the network and select the adjacent modules with the lowest information content as candidates for the next merging process.

For a network with \( L \) total LoR\textsuperscript{2}C modules, the adjacent layers \( (t^*, t^*+1) \) with the minimum combined information score can be determined as:
\begin{equation}
(t^*, t^* + 1) = \arg \min_{t \in \{1, 2, \dots, L-1\}} \text{Score}(t, t+1)
\end{equation}
where the score is defined as:
\begin{equation}
\text{Score}(t, t+1) = \text{SFS}(t) + \text{SFS}(t+1)
\end{equation}
The merging operation combines the two selected modules into a single module. The rank of the new LoR\textsuperscript{2}C module remains unchanged after the merging, and the new \( A \) and \( B \) matrices are chosen from the module with the highest information content before merging. The training and merging process is repeated until a specific number of merging operations is completed. To ensure convergence, the network is retrained for several iterations after the final merging, as illustrated in the figure.

\textbf{InjectLoR\textsuperscript{2}C:} In the injection mechanism, we dynamically evaluate the contribution of LoR\textsuperscript{2}C modules to the representation of feature information in the network, progressively reducing the number of redundant module parameters. Unlike the merging mechanism, the core idea of the injection mechanism is to replace LoR\textsuperscript{2}C modules iteratively, reducing parameter overhead while maintaining model performance. During training, the injection mechanism performs dynamic optimization over multiple iterations, replacing high-information LoR\textsuperscript{2}C modules with more efficient low-rank modules to achieve parameter efficiency.

As shown in Fig. 5, similar to the merging mechanism, we initialize and train each LoR\textsuperscript{2}C module for a certain number of epochs before calculating the SFS metrics to evaluate the information content of the LoR\textsuperscript{2}C matrices. Subsequently, in contrast to the merging mechanism, we rank all modules based on their SFS evaluation results, selecting the LoR\textsuperscript{2}C module with the lowest information content as the candidate module for replacement. This replacement process is motivated by the advantages of LoRA, which directly adjusts the Query ($W_Q$) and Value ($W_V$) weight matrices in the attention mechanism. By decomposing these matrices into low-rank components ($W = BA$), LoRA enables more fine-grained control over the attention distribution compared to modifying activation vectors.
To maintain the same parameter count, we set the rank of the LoRA module to half the rank of the original LoR\textsuperscript{2}C matrix. The injection and training steps are repeated a fixed number of times, after which the injection stops, and the model is trained for additional epochs to ensure convergence.
\subsection{Advantage Analysis}
The skip connections in LoR\textsuperscript{2}C help mitigate the gradient vanishing problem. In the model, the gradient propagation is typically calculated using the chain rule. At each layer, the presence of residual connections ensures that the input gradients not only propagate through the weight matrices of the current layer but are also directly passed to the previous layer.

Assuming the output of the \(t\)-th layer is \(h_{t+1}\), the gradient propagation formula can be written as:
\begin{equation}
\frac{\partial L}{\partial h_t} = \frac{\partial L}{\partial h_{t+1}} \frac{\partial h_{t+1}}{\partial h_t}    
\end{equation}
The output \(h_{t+1}\) at layer \(t\) consists of the contribution from the current layer's output and the input from the previous layer:
\begin{equation}
h_{t+1} = W_th_t + \text{Layer}(h_t)
\end{equation}
where \(W_t\) is the weight matrix of the current layer, and \(\text{Layer}(h_t)\) represents the non-linear transformation of the current layer. Consequently, the total gradient propagation becomes:
\begin{equation}
\frac{\partial L}{\partial h_t} = \frac{\partial L}{\partial h_{t+1}} \frac{\partial \text{Layer}(h_t)}{\partial h_t} + \frac{\partial L}{\partial h_{t+1}}B_tA_t
\end{equation}
As seen, the residual connection (or skip connection) provides a direct gradient propagation path independent of other complex non-linear transformations in the network. This ensures that even in deep networks, gradients can propagate directly through linear paths without diminishing across layers. For the low-rank matrices \(B_tA_t\) in LoR\textsuperscript{2}C, this essentially offers a more efficient and simplified linear gradient propagation path, preventing the gradients from vanishing due to excessive complex transformations.

LoR\textsuperscript{2}C, along with its optimizations ShareLoR\textsuperscript{2}C and MergeLoR\textsuperscript{2}C, reduces the number of parameters to different extents:

These reductions make LoR\textsuperscript{2}C and its variants more parameter-efficient while maintaining high performance.

\section{Experiments}
To validate the effectiveness of our proposed LoR\textsuperscript{2}C and its optimizations, we conducted extensive experiments across multiple natural language processing (NLP) tasks and models. Specifically, we evaluated natural language understanding (NLU) tasks on the GLUE dataset using the RoBERTa-base model and natural language generation (NLG) tasks on the Alpaca-Cleaned dataset using the LLAMA2-7B model. Finally, we analyzed the experimental results.
\subsection{Baselines}
We compare our methods with several Parameter-Efficient Fine-Tuning methods. 
For our evaluation, we benchmark our method against several established techniques for parameter-efficient fine-tuning. To ensure a fair and objective comparison, we reproduced the experimental configurations from previous studies and utilized their reported results. The baseline approaches selected for comparison are listed below:

\begin{itemize} \item \textbf{Full Fine-Tuning (FF)}: Fine-tuning all model parameters.

\item \textbf{AdapterD}\cite{b22} Adapters optimized for parameter efficiency.

\item \textbf{LoRA} Low-rank adaptation to reduce parameter updates.

\item \textbf{DyLoRA} Dynamic low-rank adaptation, varying rank during training.

\item \textbf{AdaLoRA} Adaptive LoRA, adjusting rank based on data.

\item \textbf{PiSSA}\cite{b23} Parameter-efficient fine-tuning via shared sparse activations.
\end{itemize}
\begin{table*}[t] 
\centering
\resizebox{0.88\textwidth}{!}{ 
\begin{tabular}{@{}lccccccc|c@{}}
\toprule
\textbf{Method} & \textbf{\#Params} & \textbf{SST-2} & \textbf{MRPC} & \textbf{CoLA} & \textbf{QNLI} & \textbf{RTE} & \textbf{STS-B} & \textbf{Avg.} \\ \midrule
FF              & 125M              & 94.8           & 90.2 & 63.6           & 92.8          & 78.7          & 91.2          & 85.2          \\ \midrule
Adpt$^D$        & 0.3M              & 94.2           & 88.5          & 60.8           & 93.1          & 71.5          & 89.7          & 83.0          \\
Adpt$^D$        & 0.9M              & 94.7           & 88.4          & 62.6           & 93.0          & 75.9          & 90.3          & 84.2          \\
LoRA            & 0.3M              & \textbf{95.1}  & 89.7          & 63.4           & \textbf{93.3} & 78.4          & \textbf{91.5} & 85.2          \\
AdaLoRA         & 0.3M              & 94.5           & 88.7          & 62.0           & 93.1          &81.0 & 90.5          & 85.0          \\
DyLoRA          & 0.3M              & 94.3           & 89.5          & 61.1           & 92.2          & 78.7          & 91.1          & 84.5          \\
PiSSA           & 0.3M              & 94.7           & 89.2          &63.8 & 92.5          & 75.5          & 90.8          & 84.4          \\
LoR\textsuperscript{2}C          & 0.15M   & 94.3& 89.2 & 62.6           & 92.3& 81.2&90.8& 85.1\\
ShareLoR\textsuperscript{2}C           & \textbf{0.075M}              & 94.1& 88.8& 62.0& 91.7& 73.4& 90.1& 83.4\\
IMLoR\textsuperscript{2}C            & $\leq 0.15$M              &94.8& \textbf{90.7}          & \textbf{64.6} & 92.6          &\textbf{82.0}           & 91.0          & \textbf{85.8}\\
\bottomrule
\end{tabular}
}
\caption{Comparison of performance across various methods on GLUE benchmark tasks. The results demonstrate accuracy or correlation for each task. IMLoR\textsuperscript{2}C in this table was configured with \( I_{\text{max}} \) and \( M_{\text{max}} \) taking their highest values from the range 0 to 6.}
\label{tab:glue_performance}
\end{table*}

\begin{table*}[t]
\centering
\resizebox{0.68\textwidth}{!}{ 
\begin{tabular}{@{}lcccccc@{}}
\toprule
\textbf{Method} & \textbf{\#Params} & \textbf{MMLU} & \textbf{BBH} & \textbf{DROP} & \textbf{HEval} & \textbf{Avg.} \\ \midrule
w/o FT         & -                 & 45.96         & 32.04        & 31.55         & 12.20          & 30.44         \\
FT             & 7B                & \textbf{47.30} & 32.72        & 29.12         & 12.80          & 30.49         \\ \midrule
LoRA           & 33.6M             & 45.64         & 32.40        & 32.46 & 15.09          & 31.40         \\
QLoRA          & 33.6M             & 45.40         & 32.81        & 28.97         & 15.24 & 30.61         \\
AdaLoRA        & 33.6M             & 45.96 & 32.85 & 31.94         & 14.02          & 31.19         \\
LoR\textsuperscript{2}C         & 16.8M    & 46.08         & 32.88 & \textbf{32.52} &15.32 & 31.7 \\
IMLoR\textsuperscript{2}C         & \textbf{12.6M}     & 46.62         & \textbf{33.00} & 32.24 & \textbf{16.46} & \textbf{32.08} \\
\bottomrule
\end{tabular}
}
\caption{Comparison of performance across different fine-tuning methods on multiple evaluation datasets.}
\label{tab:instruction_tuning}
\end{table*}
\begin{figure*}[t]
\centering
\subfigure[]
{
    \begin{minipage}[b]{.22\linewidth}
        \centering
        \includegraphics[scale=0.14]{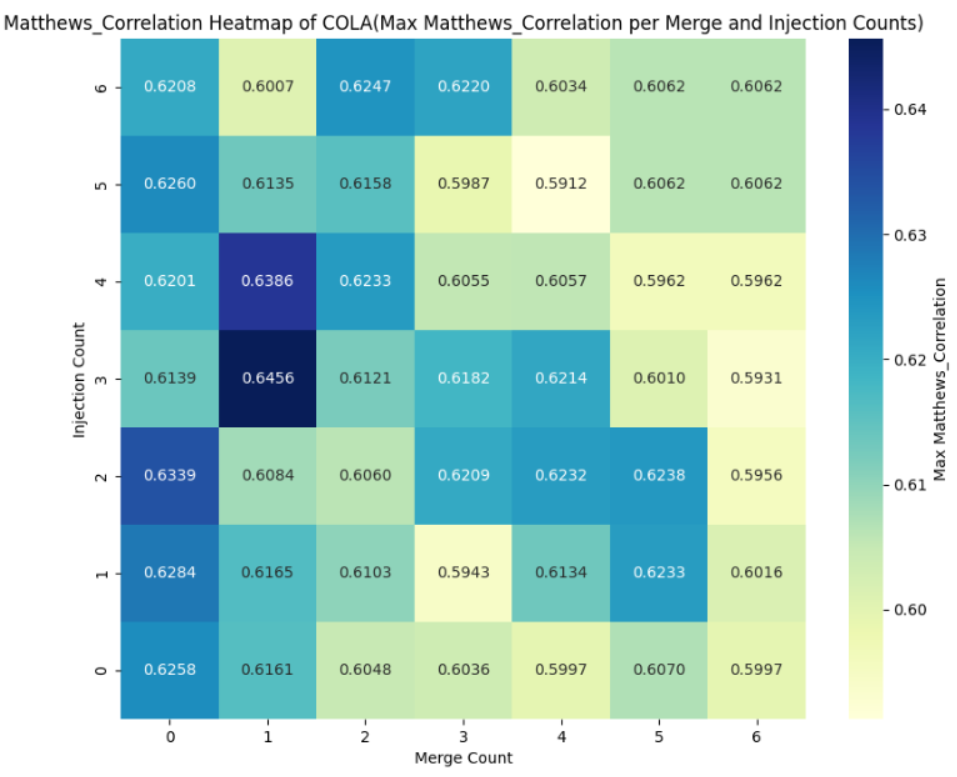}
    \end{minipage}
}
\subfigure[]
{
 	\begin{minipage}[b]{.22\linewidth}
        \centering
        \includegraphics[scale=0.14]{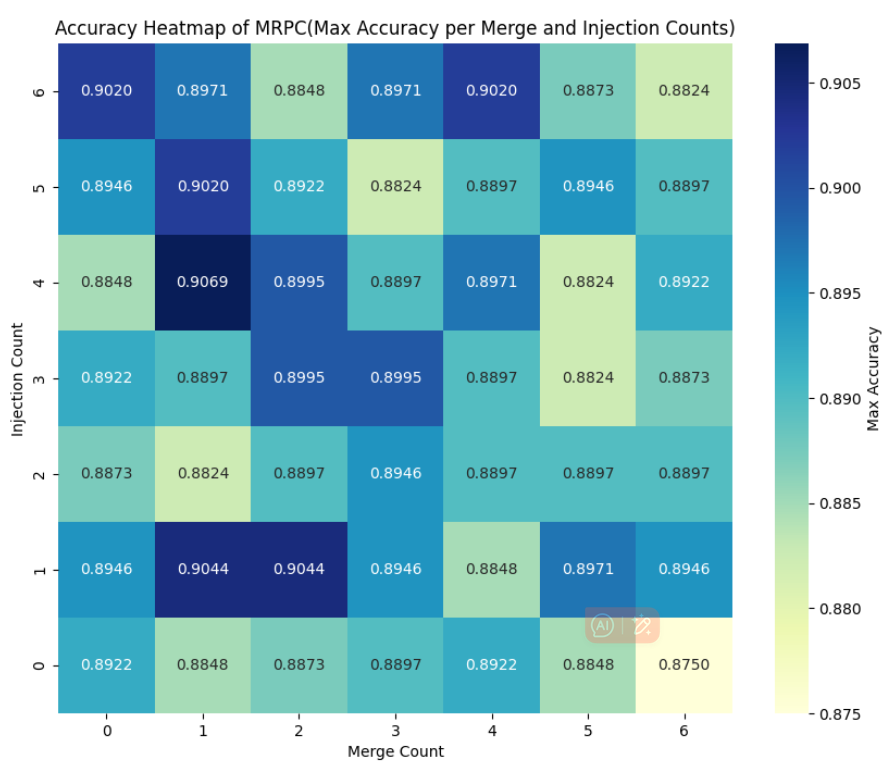}
    \end{minipage}
}
\subfigure[]
{
 	\begin{minipage}[b]{.22\linewidth}
        \centering
        \includegraphics[scale=0.14]{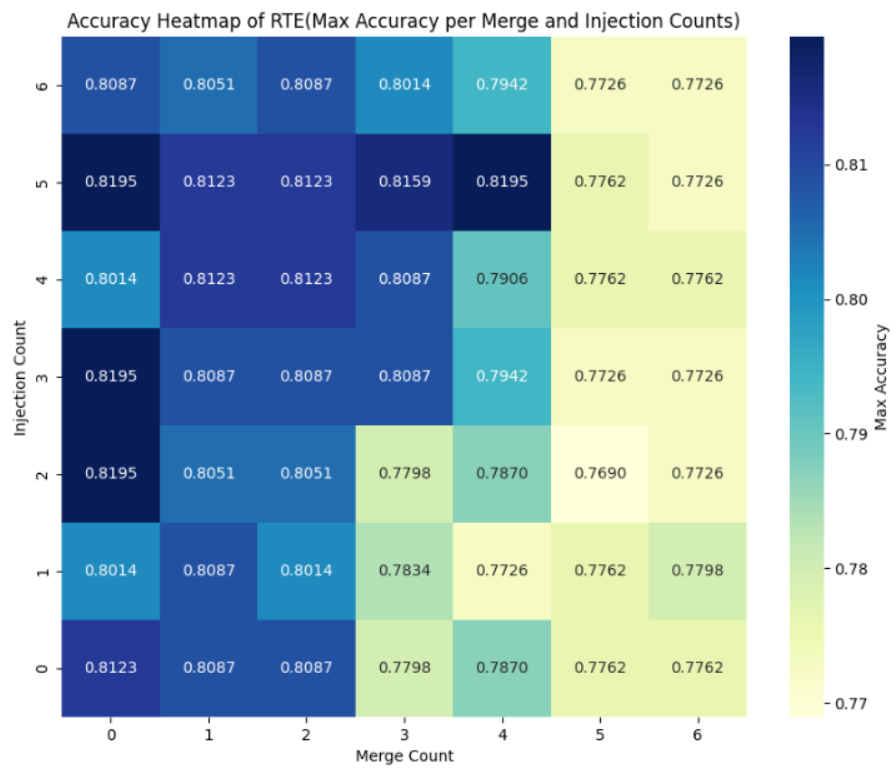}
    \end{minipage}
}\subfigure[]
{
 	\begin{minipage}[b]{.22\linewidth}
        \centering
        \includegraphics[scale=0.14]{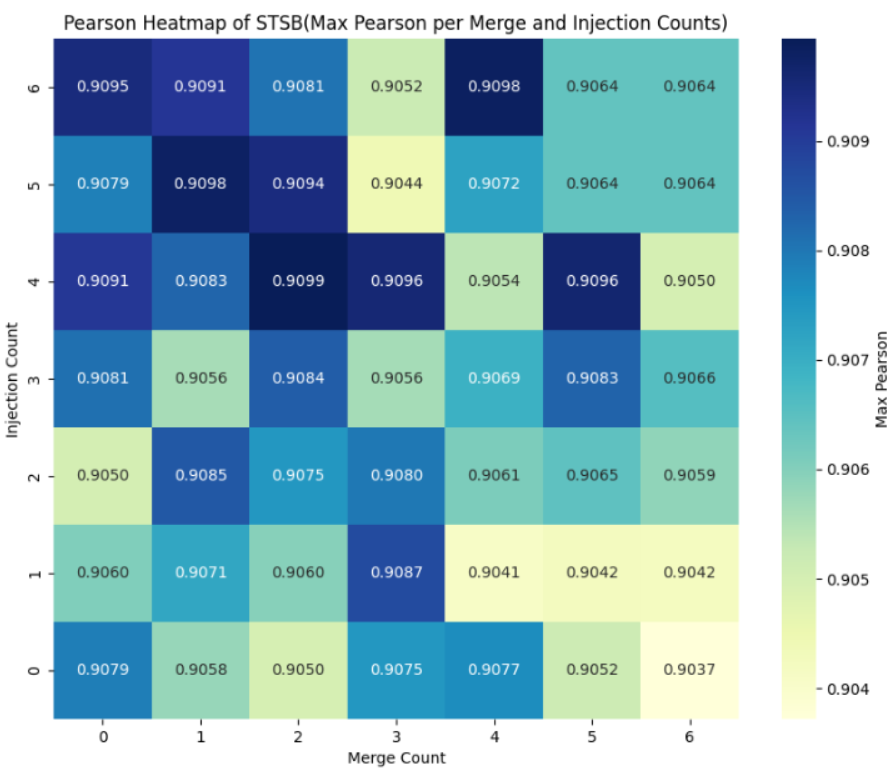}
    \end{minipage}
}
\caption{Heatmaps illustrating the performance metrics across different tasks for varying injection counts and merge counts: (a) Matthews correlation coefficient for the CoLA dataset, (b) accuracy for the MRPC dataset, (c) accuracy for the RTE dataset, and (d) Pearson correlation for the STS-B dataset. The vertical axis represents the injection count, while the horizontal axis indicates the merge count. 
}
\end{figure*}
\subsection{Natural Language Understanding}
\textbf{Dataset and Model:}
In this experiment, we used the RoBERTa-base model and the GLUE dataset. By testing on the GLUE dataset, we can objectively assess the performance of RoBERTa-base in multi-task learning and compare it with other parameter-efficient fine-tuning methods.

\textbf{Additional Details:}
For all methods, we set the rank to 8 (for the LoRA matrices in InjectLoR\textsuperscript{2}C, the rank was set to 4). For MergeLoR\textsuperscript{2}C and InjectLoR\textsuperscript{2}C, we used both methods simultaneously, which we refer to as IMLoR\textsuperscript{2}C. 

We performed merge operations and injection operations after specific intervals during training. The scheduling for these operations is determined by the following formulas:

\begin{itemize}
    \item {Merge operation scheduling:}
    \begin{equation}
    \left\lfloor \frac{\text{epoch}}{4 \times M_{\text{max}} + \epsilon} \right\rfloor + 1
     \end{equation}
    \item {Injection operation scheduling:}
    \begin{equation}
    \left\lfloor \frac{\text{epoch}}{4 \times I_{\text{max}} + \epsilon} \right\rfloor + 1
     \end{equation}
\end{itemize}

Here, $M_{\text{max}}$ represents the hyperparameter for the maximum number of merge operations, and $I_{\text{max}}$ represents the hyperparameter for the maximum number of injection operations. $\epsilon$ is defined as a small constant ($1 \times 10^{-9}$) to prevent division by zero.

In cases where both merge and injection operations are scheduled simultaneously, merge operations are prioritized. Injection operations are applied only to LoR\textsuperscript{2}C modules that have not been merged. Additionally, layers that have undergone injection operations are excluded from subsequent merge operations. Furthermore, merging between the LoR\textsuperscript{2}C modules of layers adjacent to the injected layer is prohibited.
For the GLUE benchmark, we adopted the standard settings of LoRA. For IMLoR\textsuperscript{2}C, we conducted a grid search on $M_{\text{max}}$ and $I_{\text{max}}$, with values ranging from 0 to 6, resulting in a total of $7 \times 7$ combinations, across the STS-2\cite{b24},STS-B\cite{b25},QNLI\cite{b26}, RTE\cite{b27}\cite{b28}\cite{b29}, COLA\cite{b30}, and MRPC\cite{b31} tasks. 

\subsection{Instruction Tuning}
\textbf{Dataset and Model:}
In this experiment, we also utilized the LLAMA2-7B model and the Alpaca-Cleaned dataset. For evaluation, we used INSTRUCTEVAL\cite{b33} with datasets including MMLU\cite{b34}, BBH\cite{b35}, DROP\cite{b36}, and HumanEval (HEval)\cite{b37}.

\textbf{Additional Details:}
For each task, we set different numbers of direct prompt examples: the MMLU task used 5-shot prompts, the BBH and DROP (development set) tasks used 3-shot prompts, while the HEval task did not provide any examples (0-shot prompts). The training process was carried out using the AdamW optimizer for 3 epochs, consistent with the number of training epochs of the baseline model to ensure fairness. The learning rate followed a linear adjustment schedule, starting at $3 \times 10^{-4}$, with a batch size of 128.
In the experiments on IMLoR\textsuperscript{2}C, we set both $M_{\text{max}}$ and $I_{\text{max}}$ to 3, the rank of LoR\textsuperscript{2}C to 8, and the rank of LoRA after injection to 4. Given the limited number of training epochs, we relaxed the conditions, adjusting the merging and injection operations to be performed just like Equations (12) and (13), but without rounding down.Due to computational constraints, we experimented with LoR\textsuperscript{2}C and IMLoR\textsuperscript{2}C under the condition where $M_{\text{max}}$ and $I_{\text{max}}$ are set to 8.

\subsection{Results}
The results for GLUE are summarized in Table $\mathrm{I}$. For IMLoR\textsuperscript{2}C, we use heatmaps to analyze metrics under various \( I_{\text{max}} \) and \( M_{\text{max}} \) values, with table results based on the highest metrics observed. Notably, when \( I_{\text{max}} = 0 \) and \( M_{\text{max}} = 0 \), IMLoR\textsuperscript{2}C is equivalent to LoR\textsuperscript{2}C. LoR\textsuperscript{2}C, with 0.15M parameters, ranked second among the baseline methods and excelled on RTE .ShareLoR\textsuperscript{2}C, using just 0.075M parameters, showed lower performance but maintained high efficiency under extreme constraints.IMLoR\textsuperscript{2}C achieved an average score of 85.8, outperforming all other parameter-efficient methods while using no more than 0.15M parameters.It also achieved top scores on MRPC, CoLA, and RTE. These results highlight its ability to balance reduced parameter usage with superior task performance.

Table $\mathrm{II}$ presents the evaluation results of various parameter-efficient fine-tuning methods.We compare our proposed methods.Our IMLoR\textsuperscript{2}C method achieves the best overall performance, outperforming all other methods while maintaining the lowest parameter count. Notably, IMLoR\textsuperscript{2}C surpasses the other methods in the BBH and HEval  benchmarks, demonstrating its ability to capture and generalize across diverse tasks. 
Compared to LoRA, our method achieves a comparable score while reducing parameter count by nearly {50\% . Furthermore, IMLoR\textsuperscript{2}C further improves efficiency and effectiveness while maintaining superior performance.
These results highlight the effectiveness of our proposed approach in achieving a balance between parameter efficiency and task performance, making it a promising alternative to existing fine-tuning strategies.

Fig. 6 show how different injection and merge counts impact performance. The results indicate that moderate injection counts and reasonable merge counts are critical for performance improvement. For instance, on the MRPC dataset, IMLoR\textsuperscript{2}C achieves the highest accuracy with an injection count of 4 and a merge count of 1. Similarly, on the STS-B dataset, the best Pearson correlation is obtained with both injection and merge counts set to 4.
For the RTE dataset, higher merge counts (e.g., 5) significantly enhance accuracy, highlighting the effectiveness of merging in this task. In contrast, on the CoLA dataset, the highest Matthews correlation is achieved with an injection count of 3 and a merge count of 1, further demonstrating the impact of properly configured injection and merging.
Overall, IMLoR\textsuperscript{2}C achieves outstanding performance by effectively leveraging its injection and merging mechanisms. This demonstrates that the method can significantly enhance task performance while reducing parameter usage, making it a compelling solution for parameter-efficient fine-tuning.

\subsection{Further Analyses}
\textbf{Mitigating Gradient Vanishing:} 
LoR\textsuperscript{2}C is a low-rank residual module that is parallel to the original layer, which is quite similar to the residual connection in ResNet. In the original layer, errors must be propagated through the feedforward layer and the multi-head self-attention layer. Although these modules also have residual connections, as the model depth increases, the gradient of the lower LoRA modules still experiences some vanishing. Our analysis indicates that using LoR\textsuperscript{2}C provides an alternative path for gradient propagation. Compared to traditional LoRA, which relies solely on stacked layers for backpropagation, LoR\textsuperscript{2}C introduces an additional shortcut composed of LoR\textsuperscript{2}C blocks across different layers. This additional gradient path helps suppress the gradient vanishing phenomenon in lower layers to some extent.

As shown in fig. 7, this 3D plot illustrates the gradient differences between LoR\textsuperscript{2}C fine-tuning and LoRA fine-tuning on the roberta-base model. The x-axis represents the current epoch, the y-axis represents the current layer, and the z-axis shows the ratio of the mean gradient of the LoR\textsuperscript{2}C module during backpropagation at that epoch and layer, compared to the mean gradient of the LoRA module at the same epoch and layer. From the plot, it can be observed that, for the same epoch, as the layer decreases, the ratio of the mean gradient of LoR\textsuperscript{2}C to that of LoRA increases, providing strong evidence that LoR\textsuperscript{2}C has stronger gradient propagation capability than LoRA, thus alleviating the vanishing gradient problem. Additionally, the plot reveals that the gradients of LoR\textsuperscript{2}C are generally higher than those of LoRA, which indicates that, compared to LoRA mounted on Wq and Wv, LoR\textsuperscript{2}C typically requires larger changes to achieve the same or even better performance.

\textbf{Discussion on ShareLoR\textsuperscript{2}C:} 
For traditional LoRA, various studies have explored weight-sharing mechanisms across different layers. Inspired by these, we also experimented with ShareLoR\textsuperscript{2}C to investigate whether the weight-sharing mechanisms that performed well in traditional LoRA could also be effective in LoR\textsuperscript{2}C, thereby reducing parameter count while maintaining model performance. Instead of exploring multiple adaptive sharing mechanisms, we adopted the simplest and most straightforward approach: sharing all downsampling layers across LoR\textsuperscript{2}C modules while keeping the upsampling layers independent}. This method reduces parameter count by almost half and has shown decent performance in traditional LoRA.

Unfortunately, as shown in Table I, our experimental results indicate that ShareLoR\textsuperscript{2}C performs significantly worse than LoR\textsuperscript{2}C and IMLoR\textsuperscript{2}C, failing to match or exceed their performance, unlike weight-sharing mechanisms in traditional LoRA. We believe there are two main reasons for this:

First, LoR\textsuperscript{2}C already has fewer parameters than LoRA, at least by half. Further reducing parameter count by directly sharing upsampling or downsampling layers may lead to excessively low parameter count, limiting the model’s performance upper bound.Second,Traditional LoRA is primarily used to adjust the attention mechanism, which determines what the model attends to. Although different layers have distinct roles and activation patterns, much of this variation originates from the attention mechanism. Many state-of-the-art large models have demonstrated the efficiency and power of attention mechanisms. In other words, making minor adjustments at the attention level has a similar impact on activation vectors as making more substantial adjustments at the feedforward level. Thus, despite differences between layers, fewer modifications to the attention mechanism may be sufficient, enabling lower parameter requirements and higher cross-layer similarity.

However, LoR\textsuperscript{2}C directly adjusts the activation vector, meaning the minimum information content required is higher than that of LoRA, and the similarity between different layers is lower. This discrepancy leads to the inferior performance of ShareLoR\textsuperscript{2}C. As for more heuristic and adaptive weight-sharing strategies, we leave them for future work.

\textbf{Advantages of Merging and Injection:} 
Besides LoR\textsuperscript{2}C and ShareLoR\textsuperscript{2}C, we explored another unique parameter reduction strategy: merging and injection mechanisms.
We periodically perform singular value decomposition (SVD) to assess the information content of each LoR\textsuperscript{2}C module with SFS. The metric evaluates the number of significant dimensions and their relative importance within the feature space.

Based on this evaluation, we take two key actions. First, we merge the two adjacent LoR\textsuperscript{2}C modules with the lowest information content, aiming to minimize performance loss while reducing the overall parameter count. Second, we inject a LoRA module with half the rank of LoR\textsuperscript{2}C into the attention layer of the module with the highest information content, replacing the original LoR\textsuperscript{2}C module. This process allows the model to perform more fine-grained learning and enhances its performance by refining the attention mechanism.

To ensure training stability, merging and injection are executed in separate steps within an epoch to prevent excessive structural modifications at any given time. Additionally, all merging and injection steps are completed during the first half of the total training epochs, while the latter half focuses solely on learning with a fixed structure to promote stability and convergence.
\begin{figure}[t]
    \centering
    \includegraphics[scale=0.4]{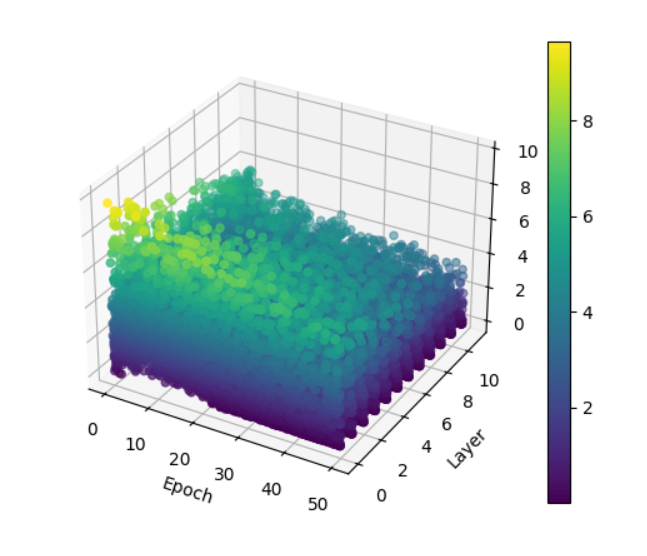} 
    \caption{3D plot of gradient ratios between LoR\textsuperscript{2}C and LoRA across epochs and layers.
}
    \label{fig:example}
\end{figure}
As shown in Table $\mathrm{I}$ and Table $\mathrm{II}$, our experimental results confirm the effectiveness of the merging and injection mechanisms. Furthermore, we observed that a small number of merging operations does not negatively impact performance and can even enhance it. This is likely because redundant parameters may interfere with model performance, occasionally leading to suboptimal decisions. Reducing these unnecessary parameters through merging can help mitigate this issue. However, excessive merging results in performance degradation, as overly coarse-grained LoR\textsuperscript{2}C adjustments across multiple layers may limit the model’s ability to effectively modify activation vectors.

Regarding injection, we identified an optimal range for its application. If too few injections are performed, the model lacks sufficient fine-grained learning capacity. On the other hand, excessive injection leads to a significant reduction in parameters, ultimately causing performance degradation.

\section{Conclusion}
In this study, we propose LoR\textsuperscript{2}C, a novel parameter-efficient fine-tuning method that reduces parameter count and mitigates the gradient vanishing problem through residual connections and low-rank matrices. We also explore optimization strategies for LoR\textsuperscript{2}C. Our method outperforms existing PEFT methods in terms of efficiency and performance across multiple tasks.
\section{Limitation}

LoR\textsuperscript{2}C introduces complex mechanisms such as parameter sharing, module merging, and dynamic injection, which increase implementation difficulty, require extensive hyperparameter tuning, and add uncertainty across tasks and models. Furthermore, the additional residual connections lead to minor inference delays, and the method's scalability to larger models or diverse tasks has yet to be validated.

Future work will focus on simplifying LoR\textsuperscript{2}C's structure to reduce implementation complexity, developing adaptive hyperparameter tuning strategies for improved automation, and expanding its applicability to larger-scale models and varied neural networks to enhance generality and practicality.

\end{document}